\documentclass[runningheads]{llncs}
\usepackage{packages}
\usepackage{caption}
\usepackage{bm}
\usepackage{hyperref}

\begin{document}
\mainmatter

\title{2D and 3D Vascular Structures Enhancement via Multiscale Fractional Anisotropy Tensor}%
\titlerunning{2D and 3D Vascular Structures Enhancement via $\overline{MFAT}$}

 \author{
     Haifa F. Alhasson \index{Alhasson,Haifa} \inst{1,2}\orcidID{0000-0002-6503-2826} \and
     Shuaa S. Alharbi \index{Alharbi,Shuaa}\inst{1,2} \orcidID{0000-0003-2121-0296}\and
     Boguslaw Obara\inst{1}\thanks{Corresponding author. Email: \email{boguslaw.obara@durham.ac.uk}}}
\authorrunning{H. Alhasson et al.}
  \institute{
      School of Computer Sciences, Durham University, Durham, UK.\\
      \and
      Computer College, Qassim University, Qassim, Kingdom of Saudi Arabia.\\}
\maketitle              
%

\begin{abstract}
	The detection of vascular structures from noisy images is a fundamental process for extracting meaningful information in many applications.  Most well-known vascular enhancing techniques often rely on Hessian-based filters. This paper investigates the feasibility and deficiencies of detecting curve-like structures using a Hessian matrix. The main contribution is a novel enhancement function, which overcomes the deficiencies of established methods.  
	Our approach has been evaluated quantitatively and qualitatively using synthetic examples and a wide range of real 2D and 3D biomedical images. Compared with other existing approaches, the experimental results prove that our proposed approach achieves high-quality curvilinear structure enhancement.
\end{abstract}


\section{Introduction} 
\label{sec:intro}
The enhancement of vessel-like structures in images plays an important role in various applications of computer vision, image processing, and medical analysis. 
The enhancement phase can be immediately improved upon by advancing the acquisition, interpretation and image analysis techniques. 
A wide range of curvilinear structures enhancement methods analyse image derivatives, most of them employ the analysis of Hessian matrix, such as~\cite{frangi1998multiscale,sato1998three,meijering2004design}. The Hessian matrix is based on second-order Gaussian derivatives, calculated at different scales (controlled by standard deviation $\sigma$). This enables the differentiation between particular shapes, such as rounded, tubular, and planar structures. These approaches suffer from many deficiencies, which can be seen in different curve patterns and in the suppression of junctions and rounded structures~\cite{jerman2016enhancement}.

Recently, the use of diffusion tensors (such as the Regularised Volume Ratio tensor (RVR)\cite{jerman2016enhancement}) has improved the detection of vessel-like structures. 
The Fractional Anisotropic Tensor (FAT) is another well-known diffusion tensor measure, which has been reviewed in~\cite{peeters2009analysis}. FAT measures the variance of eigenvalues across different structures, i.e., it measures the change of anisotropy along the vessels. 
In terms of ellipsoid glyphs, cigar-shaped (linear) and pancake-shaped ellipsoids (planar) can result in equal FAT measures though their shapes differ greatly~\cite{hansen2011visualization}. This feature gives FAT less potential to fail in the junctions and has more chance of getting closer to uniformity in the final response. Thus, FAT plays a major role in many attempts towards diffusion tensor regularization~\cite{coulon2004diffusion}.
An optimal enhancement function should achieve a high and uniform response to i) variable vascular morphology, ii) the intensity non-uniformities caused by blood contrast or in the background, iii) unambiguity of the vessel boundary, and iv) background noise.

Our contribution in this paper is a novel multiscale approach for either 2D or 3D images by preprocessing the eigenvalues and junctions reconstructing at each scale. Our proposed method has shown a surpassed performance when compared to the competing state-of-the-art approaches.


\section{Related Work} 

One of the early attempts to use eigenvalue analysis for vessel enhancement was by Sato~\cite{sato1998three}. A later attempt that received wide acceptance was in the work of Frangi et al.~\cite{frangi1998multiscale}. In their work, they proposed a Hessian-based approach, known as Vesselness, to enhance curve-like features.
The Vesselness measure is used to describe an image whenever a dark curvilinear structure appears with respect to the background.
However, the main drawback of this approach are the very small curve-like feature responses at junctions, due to the large eigenvalues.
The use of the Hessian matrix was further developed in the Neuriteness method, proposed by Meijering et al.~\cite{meijering2004design}. 
It consists of a detection stage, which implies a feasible neurite with a value in every individual pixel of the image, and the actual tracing stage.  
This last stage determines which pixels are successive and which are most likely to reflect the centerline of the neurites.
Furthermore, Obara et al.~\cite{obara2012contrast} used a Phase Congruency Tensor (PCT) by~\cite{kovesi2003phase} in combination with Vesselness (PCT ves.)~\cite{frangi1998multiscale} and Neuriteness (PCT neu.)~\cite{meijering2004design} in order to improve their results in detecting edges and to not rely on image intensities. The advantage of this approach is its insensitivity to intensity and noise variations in images. 

Recently, motivated by the detection of spherical diffusion tensors, Jerman et al.~\cite{jerman2016enhancement} proposed a new measure based on the Volume Ratio, which overcomes the deficiencies of using a Hessian matrix such as: non-uniformity, variation of eigenvalues with image intensity, and non-uniformity of enhancement across scales. 
However, the problem of having lower intensities on junctions and crossings has not been completely solved by this approach.
For more details, we provide further background of the enhancement approaches (please refer to the supplementary materials).


\section{Methodology}

In this section, we introduce a novel approach in order to enhance vessel-like structure in images.
We hypothesize that an enhancement function should take the degree of anisotropy of the target structure into account, should be preserving the transactions between isotropic and anisotropic tissues and should be robust to sustain low-magnitude eigenvalues. Furthermore, an enhancement function should solve the fitting problem in the elliptical cross-section structures, which yield a uniform response across different vascular structures and more effective suppression of background noise without affecting the junctions or cross-sections.
 
Biological tissue samples are often anisotropic, because the cell and vessel membranes limit the motion of water molecules.
Since the shape of the curvilinear structures is anisotropic, the junctions in these tissues have an isotropic shape, which explains why other established methods could not detect non-circular cross-sections~\cite{frangi1998multiscale,meijering2004design} or having less uniformity in the junctions and crossing~\cite{jerman2016enhancement}.
In order to avoid the low filter response at junctions, we regularise eigenvalues, calculated from Hessian representation of an image $I(\bm{x})$ at each scale $\sigma$, to fulfill the following condition for 3D images:
\begin{equation}\label{eq:lambda}
 {\lambda}_2\geq{\lambda}_3 \wedge |{\lambda}_{2,3}|\gg |{\lambda}_{1}|.
\end{equation}
This eigenvalues regularization process is combined with a junctions reconstruction step at each scale $\sigma$. This paper proposed a new Hessian-based enhancement approach called Multiscale-Fractional Anisotropic Tensor $\overline{MFAT}$.
\subsection{Fractional Anisotropic Tensor-based Vascular structures Enhancement}
\subsubsection{Enhancement function in 3D}
The anisotropy on a voxel level is quantified in terms of FAT and is expressed as an invariant of the three independent diffusion tensor eigenvalues at each scale $\sigma$, and it is expressed in~\cite{hansen2011visualization} as: 
\begin{equation}
\resizebox{0.46\linewidth}{!}{$
 \label{eq:FA}
{FAT}^{\sigma}_{\lambda} = \sqrt{\frac{3}{2}} \sqrt{\frac{{{(\lambda_1- \overline{D}_{\lambda})^2}+{(\lambda_2- \overline{D}_{\lambda})^2}+{(\lambda_3- \overline{D}_{\lambda})^2}}}{{\lambda_1}^2+{\lambda_2}^2+{\lambda_3}^2}}$}.
\end{equation}
The response of $FAT^{\sigma}_{\lambda}$ ranges between 0 and 1.
The mean diffusivity $\overline{D}_{\lambda}$ is defined as:
\begin{equation}\label{eq:mean}
\overline{D}_{\lambda} = \frac{T_r}{3},
\end{equation}
where $T_r$ refers to the trace of diffusion tensor, which represents the affected area much more accurately than images, representing the diffusion in only one direction, and defined as:
\begin{equation}\label{eq:tr}
T_r = \sum_{i=1}^{3}\lambda_{i}.
\end{equation}

Recently, Pardos et al.~\cite{prados2010analysis} proposed another representation of anisotropic diffusion tensors, in probabilistic form based on a ternary diagram, called \textit{Finetti Diagram}~\cite{aitchison1986statistical}, analysed for each tensor, and proved its feasibility. 
They describe the main limitation of eigenvalue-based measures as its partial representation of the tensor information which is only related to image intensities.
They proved that the probability-based FAT measure has better detection of curve-like structures. 
The Probabilistic Fractional Anisotropic Tensor $FAT_{p}$, at each scale $\sigma$, can be expressed as:
  \begin{equation}
  \resizebox{0.46\linewidth}{!}{$
  \label{eq:PFAT}
  {FAT}^{\sigma}_{p} = \sqrt{\frac{3}{2}} \sqrt{\frac{{{(p_1- \overline{D}_p)^2}+{(p_2- \overline{D}_p)^2}+{(p_3- \overline{D}_p)^2}}}{{p_1}^2+{p_2}^2+{p_3}^2}}$}.
  \end{equation}
 The relative importance of each ellipsoid axis $p_i$ is defined as:
\begin{equation}
p_i = \frac{\lambda_{ i }}{T_r},
\label{eq:importance}
\end{equation}
where the mean of diffusivity is set to be $\overline{D}_p = \frac{1}{3}$ as in~\cite{prados2010analysis}.

\par Both forms of fractional anisotropic have been modified in this work from the original version in Equations \ref{eq:FA} and \ref{eq:PFAT}. Our enhancement function is based on the modifications that have been done in~\cite{jerman2016enhancement}. They add absolute values to account for differently signed eigenvalues, which results in a more uniform response. Also, they eliminate $\lambda_{ 1 }$ to get more normalized results, and by regularizing the value of $\lambda_{ 3 }$ at each scale $\sigma$ using cut-off threshold $\tau$.
Our new enhancement function of $\overline{FAT}^{\sigma}_{\lambda}$ is defined as follows:
\begin{equation}
\label{eq:FATnew}
\resizebox{0.46\linewidth}{!}{$
\overline{FAT}^{\sigma}_{\lambda} = \sqrt{\frac{3}{2}} \sqrt{\frac{{{(\lambda_2- \overline{D}_{\lambda})^2}+{(\lambda_{\rho})- \overline{D}_{\lambda})^2}+{(\lambda_{\nu}- \overline{D}_{\lambda})^2}}}{{\lambda_2}^2+{\lambda_{\rho}}^2+{\lambda_{\nu}}^2}}$},
\end{equation}
where $\tau_\rho$ is corresponding to $\lambda_{ \rho }$, which is adopted from the work has been done in~\cite{jerman2016enhancement}. We propose using another cut-off thresholding $\tau_\nu$ to regulate $\lambda_{ 3 }$ at each point $\bm{x}$ in each scale $\sigma$ to fulfil the condition in Equation~\ref{eq:lambda}. Both $\lambda_{ \rho}$ and $\lambda_{\nu}$ can be obtained from:
\begin{equation}
\label{eq:newlambda}
\resizebox{0.7\hsize}{!}{$%
\lambda_{ \rho,\nu} = \begin{cases} \lambda_{3}(\bm{x},\sigma) \quad & $if$\quad \lambda_{ 3 }(\bm{x},\sigma) > \tau_{ \rho,\nu} \max_{\bm{x}}\big(\lambda_{ 3 }(\bm{x},\sigma)\big),\\
\tau_{ \rho,\nu} \max_{\bm{x}}\big(\lambda_{ 3 }(\bm{x},\sigma)\big) \quad & $if$\quad 0 < \lambda_{ 3 }(\bm{x},\sigma) \leqslant \tau_{ \rho,\nu} {\max}_{\bm{x}}\big(\lambda_{ 3 }(\bm{x},\sigma)\big),\\
0\quad & otherwise, \end{cases}$}
\end{equation} 
 where $\tau_\rho$ and $\tau_\nu$ are between $[0,1]$.
With the above eigenvalues regularization, both Equations \ref{eq:PFAT} and \ref{eq:importance} can be written as follows:
\begin{equation}
\label{eq:PFATnew}
\resizebox{0.46\linewidth}{!}{$
	\overline{FAT}^{\sigma}_{p} = \sqrt{\frac{3}{2}} \sqrt{\frac{{{(p_2- \overline{D}_p)^2}+{(p_{\rho}- \overline{D}_p)^2}+{(p_{\nu}- \overline{D}_p)^2}}}{{p_2}^2+{p_{\rho}}^2+{p_{\nu}}^2}},$}
\end{equation}
where 
\begin{equation}\nonumber
\resizebox{0.4\linewidth}{!}{$
p_{2}= \abs{\frac{\lambda_{2}}{T_r}}, \quad
p_{\rho} = \abs{\frac{\lambda_{\rho}}{T_r}}, \quad
p_{\nu} = \abs{\frac{\lambda_{\nu}}{T_r}} $}. 
\end{equation}
The inverted response of either Equation~\ref{eq:FATnew} or Equation~\ref{eq:PFATnew} assure a positive response at vessel and the junctions. 
Furthermore, in order to remove noise from the background, we add more restrictions. 
Therefore, the response $R^{\sigma}_{\lambda,p}$ can be written as follows:
\begin{equation}
\resizebox{0.55\hsize}{!}{$%
	R^{\sigma}_{\lambda,p}= \begin{cases} 0 \quad & $if$\quad  \lambda_{\rho} > \lambda_{\rho} - \lambda_{2} \vee \lambda_{\rho} \geq 0 \vee \lambda_{2} \geq 0,  \\
	1 \quad & $if$\quad  \lambda_{\rho} - \lambda_{2} = \max_{\bm{x}}(\lambda_{\rho} - \lambda_{2}), \\
	 {1-\overline{FAT}^{\sigma}_{\lambda,p}} & otherwise. \end{cases}$}
\end{equation}
Using the similar concept of the magnitude regularization in~\cite{coulon2004diffusion}, the junctions reconstruction is obtained by a maximized co-addition of response at each scale $\sigma$ and the final enhancement function 
$\overline{MFAT}_{\lambda,p}$ using either Equation~\ref{eq:FATnew} or ~\ref{eq:PFATnew} as follows:
\begin{equation}
\resizebox{0.46\linewidth}{!}{$
		\overline{MFAT}^{\sigma}_{\lambda,p} = \overline{MFAT}^{\sigma-1}_{\lambda,p} + \delta \tanh \big( R^{\sigma}_{\lambda,p} - \delta \big)$}, 
\end{equation}
\begin{equation}\label{eq:MFAT}
\resizebox{0.37\linewidth}{!}{$
	\hskip-23pt \overline{MFAT}_{\lambda,p} = \max_{\sigma} \bigg( \overline{MFAT}^{\sigma}_{\lambda,p},R^{\sigma}_{\lambda,p} \bigg) $},
\end{equation}
where $\sigma$ is the current scale and $\sigma-1$ is a previous scale. $\delta$ is the step size during the calculation of the solution. 
Considered as possible improvements at the beginning of this section, our enhancement method produce a highly uniform response that is very close to the ground truth of typical curvilinear structures. 
\subsubsection{Enhancement in 2D}
Our proposed function $\overline{MFAT}_{\lambda,p}$ can be also defined for 2D case. In such case, there are three eigenvalues $\lambda_{ 2 }, \lambda_{ \rho }$, and $\lambda_{\nu}$ that are defined in Equation~\ref{eq:newlambda}. The corresponding response $R^{\sigma}_{\lambda,p}$ for 2D images as follows:
\begin{equation}
\resizebox{0.85\hsize}{!}{$%
	R^{\sigma}_{\lambda,p}= \begin{cases} 0 \quad & $if$\quad  \lambda_{\rho} > \lambda_{\rho} - \lambda_{2}  \vee \lambda_{\rho} \geq 0 \vee \lambda_{2} \geq 0 \vee \lambda_{\rho} - \lambda_{2} < \max_{\bm{x}}(\lambda_{\rho} - \lambda_{2}),\\
	1 \quad & $if$\quad  \lambda_{\rho} - \lambda_{2} = \min_{\bm{x}}(\lambda_{\rho} - \lambda_{2}), \\
	{1-\overline{FAT}^{\sigma}_{\lambda,p}} & otherwise. \end{cases}$}
\end{equation}

\section{Results} \label{sec:result}
In this section, we present quantitative and qualitative validations for our proposed approach against both synthetic and real-world 2D and 3D imaging data. 
We then compare the results with state-of-the-art approaches. 
The Receiver Operating Characteristic (ROC) curve~\cite{fawcett2006introduction} is widely adopted in similar analysis. We used the Area Under the Curve (AUC) of the ROC curve  to compare the curvilinear structure enhancement approaches.
\subsection{Profile Analysis}
\begin{figure}[!ht]
	\centering
	\vskip-5pt
	\hskip-1 cm
	\begin{subfigure}{0.17\linewidth}\centering
		\begin{tikzpicture}[scale=1]
    	\node(0,0){\includegraphics[width=0.7\linewidth,trim=0 10mm 0 10mm,clip]{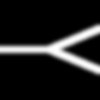}};
	    \draw[dashed, line width=1.5pt, red] (0,0.5) -- (0,-0.5);
 	   \end{tikzpicture}
 	\end{subfigure}
 \hskip 3.67 cm
 \begin{subfigure}{0.17\linewidth}\centering
 	\begin{tikzpicture}[scale=1]
 	\node(0,0){\includegraphics[width=0.7\linewidth,trim=0 10mm 0 10mm,clip]{images/SyntheticCurveNo}};
 	\draw[dashed, line width=1.5pt, red] (1,0) -- (-1,0);
 	\end{tikzpicture}
 \end{subfigure}\\
	\begin{subfigure}{0.45\linewidth}\centering
		\begin{tikzpicture}
		\begin{axis}[
		width = \linewidth,
		xlabel={Distance from Vessel Centreline},
		ylabel={Normalised Response},
		label style={font=\tiny},
		tick label style={font=\footnotesize},
		ytickmin=0, ytickmax=1,
		ymin=0, ymax=1,
		xmin=-10, xmax=10,
		enlargelimits=false,
		legend entries = {Raw image, Vesselness, Neuriteness, PCT ves., PCT neu., RVR, $\overline{MFAT}_{\lambda}$(Ours),$\overline{MFAT}_{p}$(Ours)},
		legend columns = 1,
		legend style={font=\tiny, line width=1pt, draw=none,},
		legend to name=leg,
		grid=major, 
		grid style={dashed,gray!30}, 
		]
		\addplot[color=black,loosely dashed] table [x index=0, y index=1, col sep=comma] {images/datFiles/profiles.dat};
		\addplot[color=brewerDark1] table [x index=0, y index=4, col sep=comma] {images/datFiles/profiles.dat};
		\addplot[color=brewerDark2] table [x index=0, y index=5, col sep=comma] {images/datFiles/profiles.dat};
		\addplot[color=brewerDark3] table [x index=0, y index=6, col sep=comma] {images/datFiles/profiles.dat};
		\addplot[color=brewerDark4] table [x index=0, y index=7, col sep=comma] {images/datFiles/profiles.dat};
		\addplot[color=brewerDark5] table [x index=0, y index=8, col sep=comma] {images/datFiles/profiles.dat};
		\addplot[color=red] table [x index=0, y index=2, col sep=comma] {images/datFiles/profiles.dat};
		\addplot[color=black] table [x index=0, y index=3, col sep=comma] {images/datFiles/profiles.dat};
		\end{axis}
		\end{tikzpicture}
	\end{subfigure}
	\begin{subfigure}{0.45\linewidth}\centering
		\begin{tikzpicture}
		\begin{axis}[
		width = \linewidth,
		xlabel={Distance from Vessel Centreline},
		ylabel={Normalised Response},
		label style={font=\tiny},
		tick label style={font=\footnotesize},
		ytickmin=0, ytickmax=1,
		xmin=-35, xmax=35,
		ymin=0, ymax=1,
		enlargelimits=false,
		grid=major, 
		grid style={dashed,gray!30}, 
		]
		\addplot[color=black,loosely dashed] table [x index=0, y index=1, col sep=comma] {images/datFiles/profiles2.dat};
		\addplot[color=brewerDark1] table [x index=0, y index=4, col sep=comma] {images/datFiles/profiles2.dat};
		\addplot[color=brewerDark2] table [x index=0, y index=5, col sep=comma] {images/datFiles/profiles2.dat};
		\addplot[color=brewerDark3] table [x index=0, y index=6, col sep=comma] {images/datFiles/profiles2.dat};
		\addplot[color=brewerDark4] table [x index=0, y index=7, col sep=comma] {images/datFiles/profiles2.dat};
		\addplot[color=brewerDark5] table [x index=0, y index=8, col sep=comma] {images/datFiles/profiles2.dat};
		\addplot[color=red] table [x index=0, y index=2, col sep=comma] {images/datFiles/profiles2.dat};
		\addplot[color=black] table [x index=0, y index=3, col sep=comma] {images/datFiles/profiles2.dat};
		\end{axis}
		\end{tikzpicture}
	\end{subfigure}
	\hskip-7pt
	\begin{subfigure}{0.1\linewidth}\centering
		\vskip-15pt
		\pgfplotslegendfromname{leg}
	\end{subfigure}
	\caption{Cross-sectional profile of a synthetic vessel image (black, dashed line), non-crossing junction in vessel-like structure enhanced by the proposed $\overline{MFAT}_{\lambda,p}$ methods (black and red solid line) and by the state-of-the-art methods (see legend for colours). All images were normalised such that the brightest pixel in the whole image has a value of $1$ and the darkest a value of $0$.}\label{fig:profile}
\end{figure}

The profile of our proposed $\overline{MFAT}_{\lambda,p}$ methods and other state-of-the-art enhancement methods on a synthetic, vessel-like structure are shown in Fig.~\ref{fig:profile}. We evaluate our approach using Equations~\ref{eq:FATnew} and~\ref{eq:PFATnew}, which refer them to $\overline{MFAT}_{\lambda} $ and $\overline{MFAT}_{p}$, respectively. 
Hessian-based methods, such as vesselness and neuritness, have an enhanced signal at the center of the vessel, i.e., a peak value of one at the centre-line of the vessel, and their respective value quickly drops off and decreases with the perceived thickness of the vessel. 
On the other hand, the most recent approach (RVR), despite producing a defined response, still shows a poor response to non-crossing junctions. The proposed approach matches all the features of previous methods and shows a more uniform response at non-crossing and crossing junctions.

\subsection{Application to 2D Retinal Images}
Although a visual inspection can provide some qualitative information, a more rigorous form of quantitative validation is required to measure the effectiveness of curvilinear structure enhancement approaches.  
The quality of the approach is measured by using the following publicly available retinal image datasets:  DRIVE~\cite{niemeijer2004comparative}, STARE~\cite{hoover2000locating}, and HRF~\cite{odstrcilik2013retinal}. 
In particular, we evaluate our approach alongside state-of-the-art approaches, calculating the mean ROC curve and the mean of AUC between the enhanced images and the ground truth. 
The results are shown in Fig.~\ref{fig:rtina}, Fig.~\ref{fig:retinaROC}, and Table~\ref{tab:auc2} (results for DRIVE, STARE and unhealthy HRF datasets can be found in the supplementary material). 
A higher AUC value indicates a better enhancement of curvilinear structures, with a value of $1$ indicating that the enhanced image is identical to the ground truth image. 
Our experimental results clearly show that the proposed approaches outperform state-of-the-art approaches, as illustrated as mean AUC in Table~\ref{tab:auc2}. 
\newcommand{\cWidth}{0.236}
\begin{figure*}[!h]
	\centering
	\vskip-5pt
	\begin{subfigure}[t]{\cWidth\linewidth}\includegraphics[width=\linewidth]{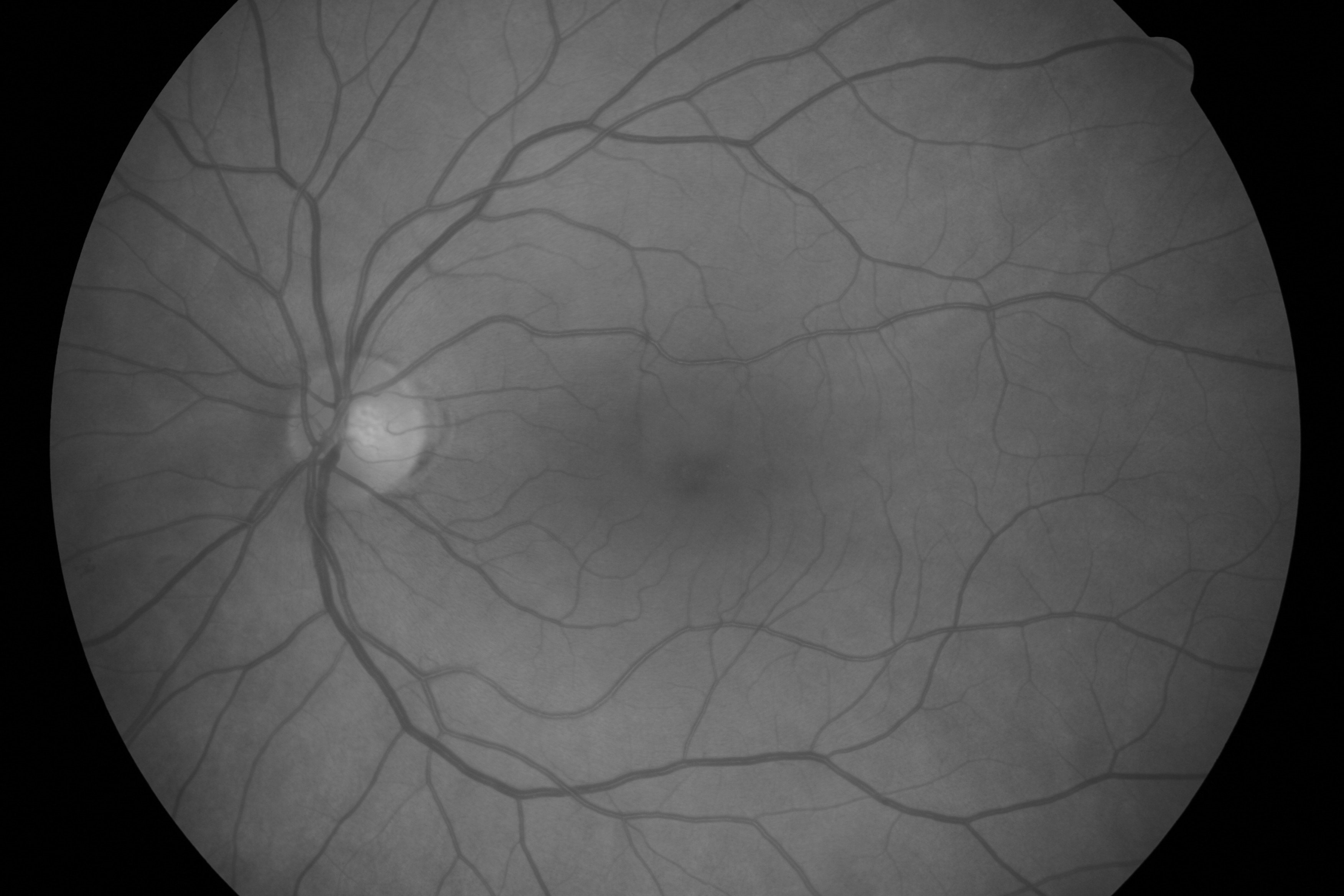}
		\caption{\quad}
	\end{subfigure}
	\begin{subfigure}[t]{\cWidth\linewidth}\includegraphics[width=\linewidth]{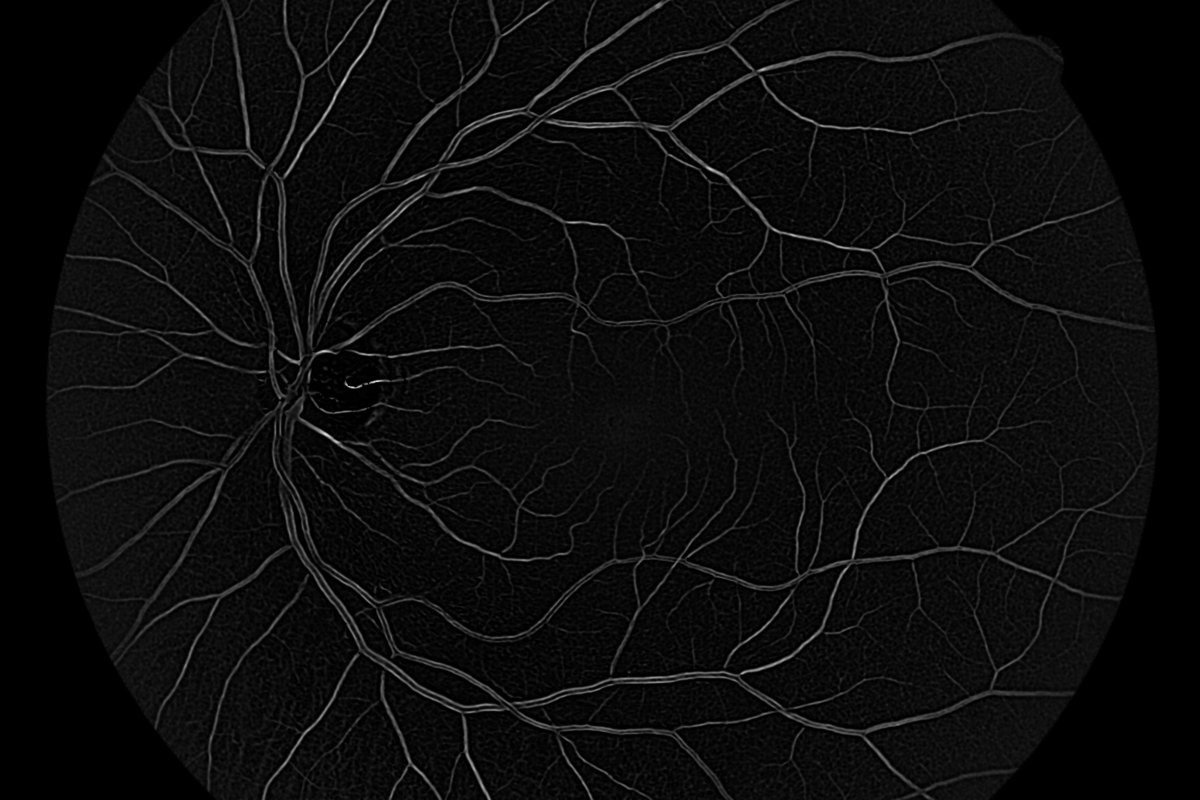}
		\caption{\quad}
	\end{subfigure}
	\begin{subfigure}[t]{\cWidth\linewidth}\includegraphics[width=\linewidth]{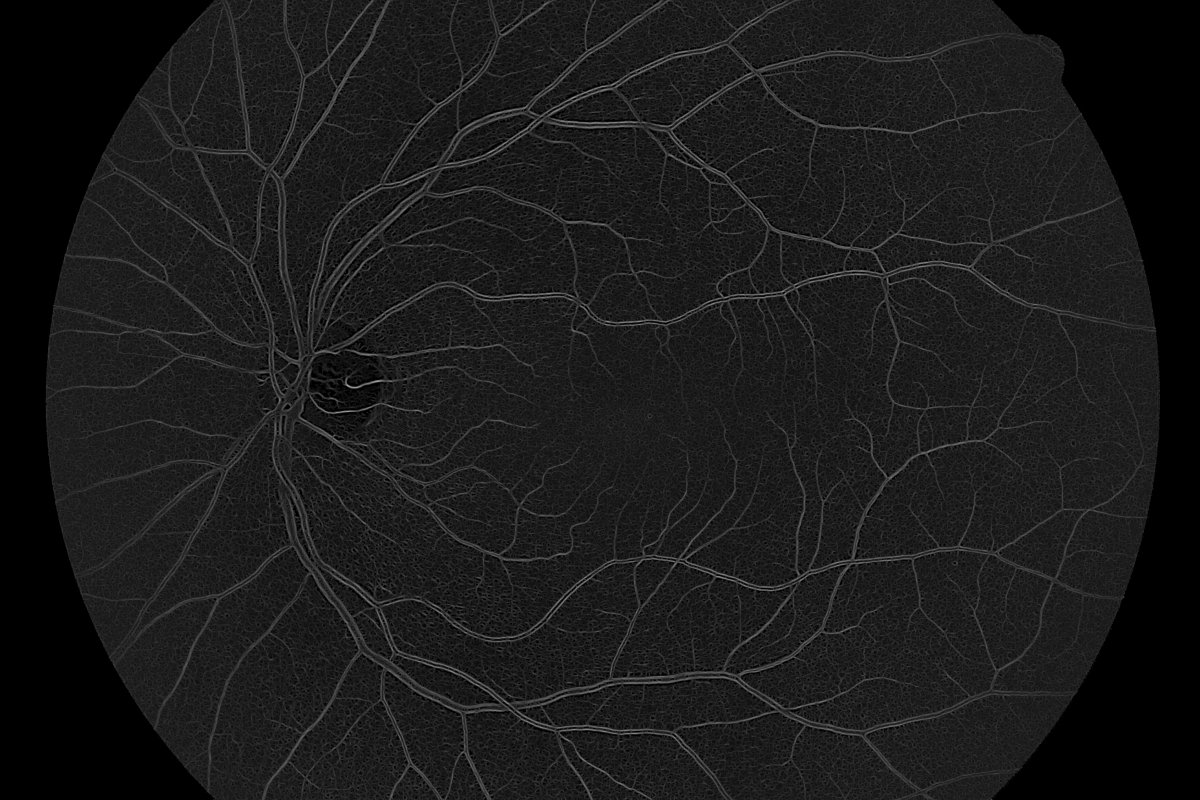}
		\caption{\quad}
	\end{subfigure}
	\begin{subfigure}[t]{\cWidth\linewidth}\includegraphics[width=\linewidth]{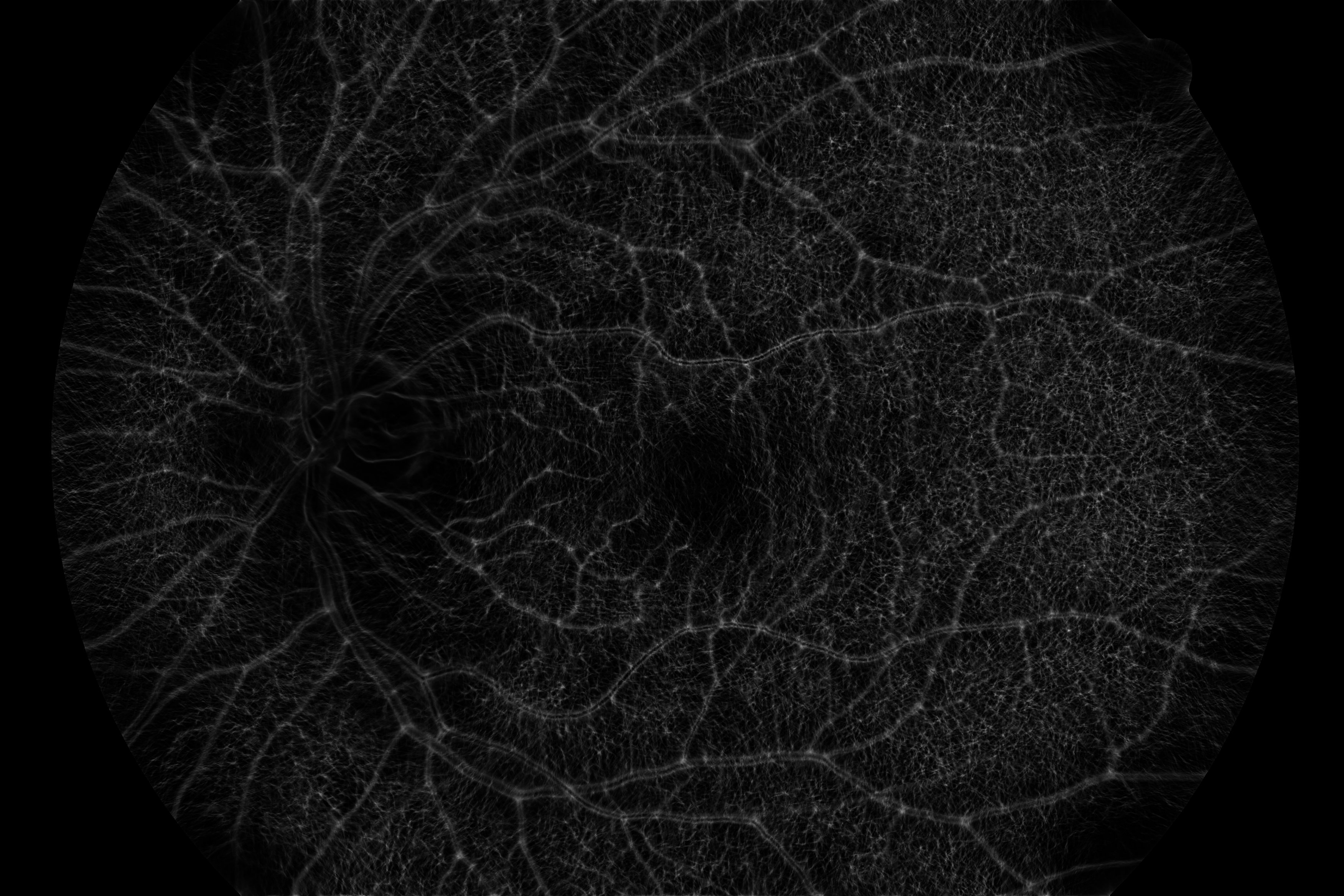}
		\caption{\quad}
	\end{subfigure}
    \\
	\begin{subfigure}[t]{\cWidth\linewidth}\includegraphics[width=\linewidth]{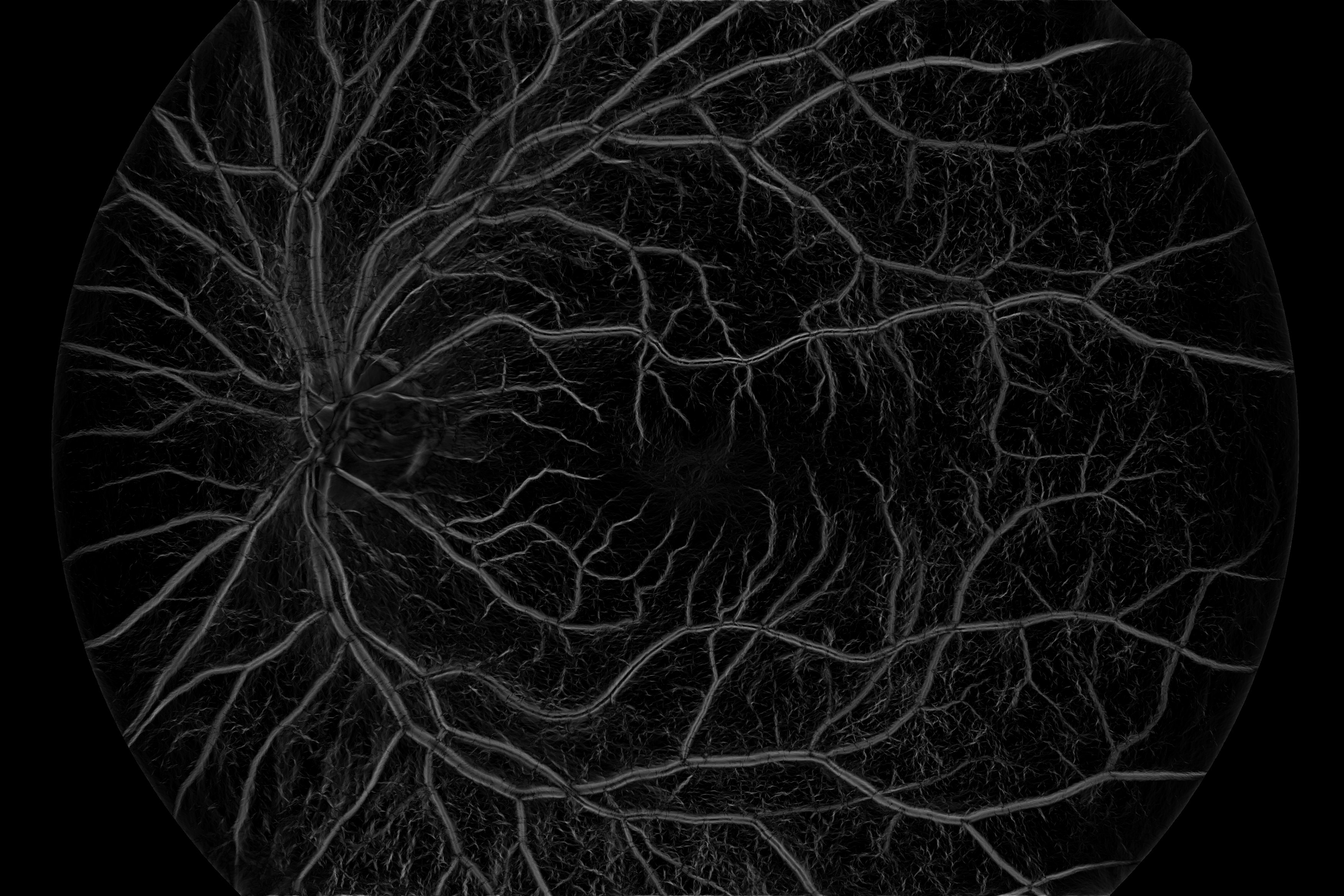}
		\caption{\quad}
	\end{subfigure}
	\begin{subfigure}[t]{\cWidth\linewidth}\includegraphics[width=\linewidth]{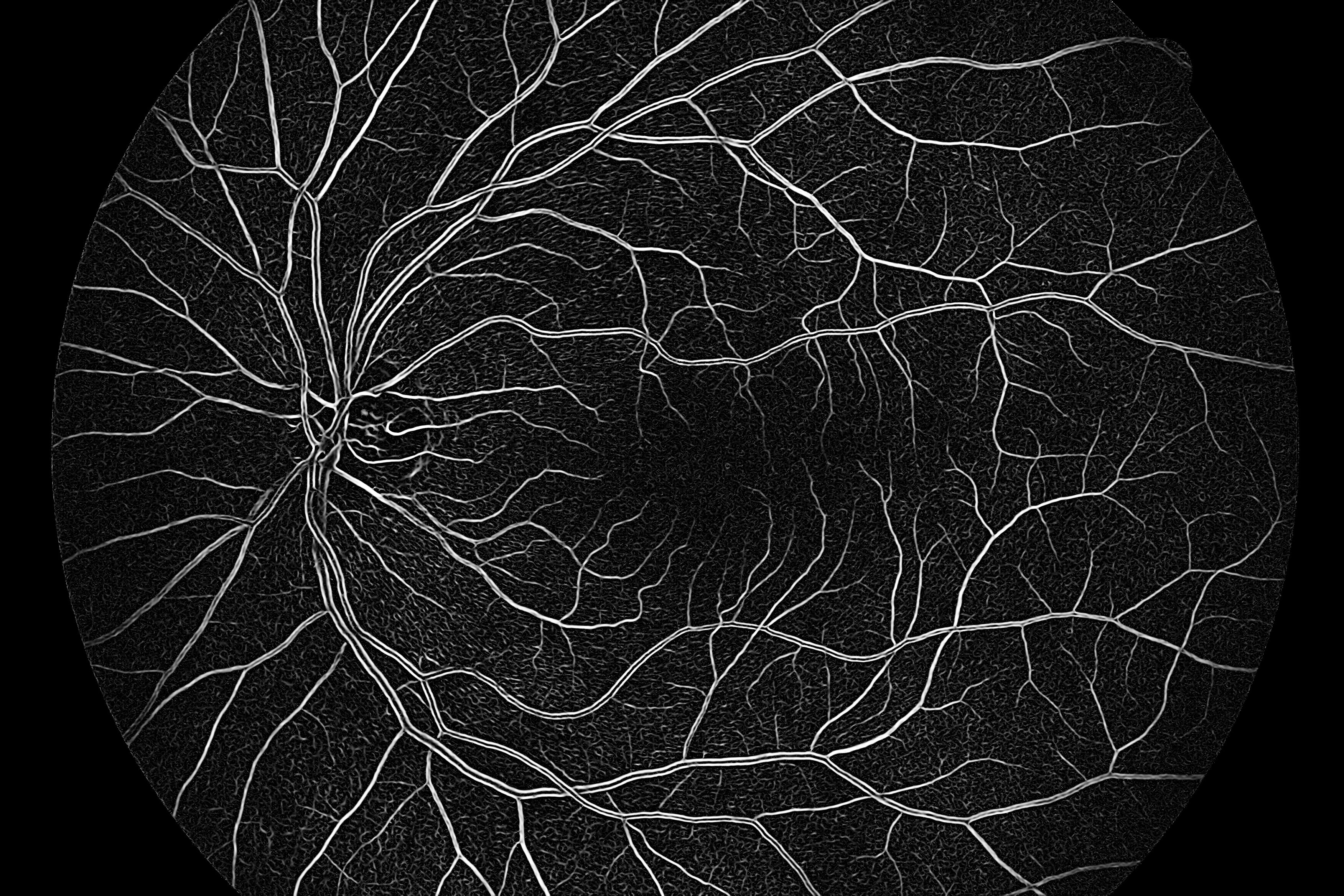}
		\caption{\quad}
	\end{subfigure}
	\begin{subfigure}[t]{\cWidth\linewidth}\includegraphics[width=\linewidth]{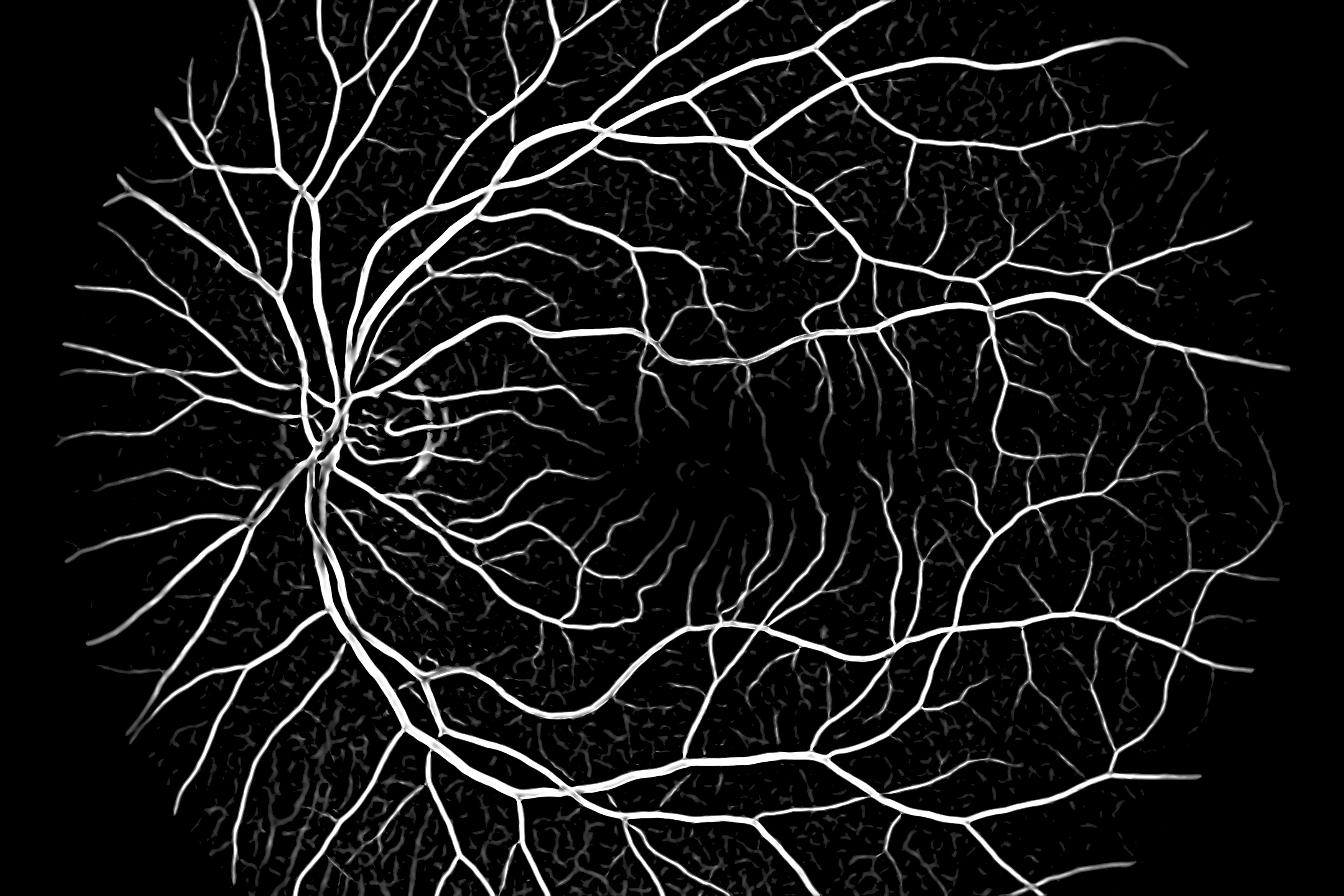}
		\caption{(Ours)}
	\end{subfigure}
	\begin{subfigure}[t]{\cWidth\linewidth}\includegraphics[width=\linewidth]{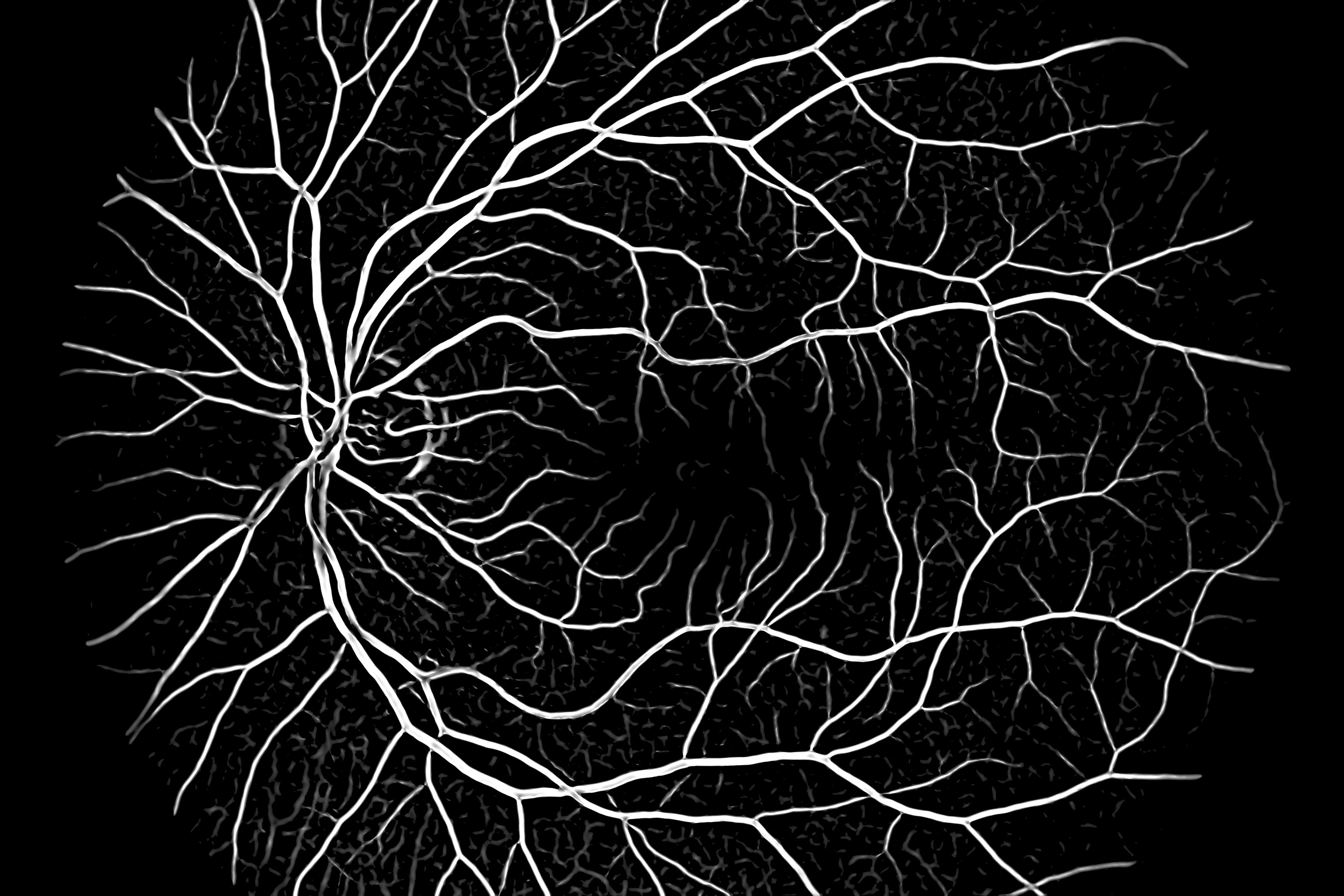}
	    \caption{(Ours)}
    \end{subfigure}
	\caption{A sample image from the healthy HRF retina dataset, alongside the enhanced images from the state-of-the-art approaches. (a) The original grayscale image, (b) Vesselness~\cite{frangi1998multiscale}, (c) Neuriteness~\cite{meijering2004design}, (d) PCT ves.~\cite{obara2012contrast}, (e) PCT neu.~\cite{obara2012contrast}, (f) RVR~\cite{jerman2016enhancement}, (g) $\overline{MFAT}_{\lambda}$ and (h) $\overline{MFAT}_{p}$ methods.}
	\label{fig:rtina}
\end{figure*}

\newcommand{\sfigure}{0.32}
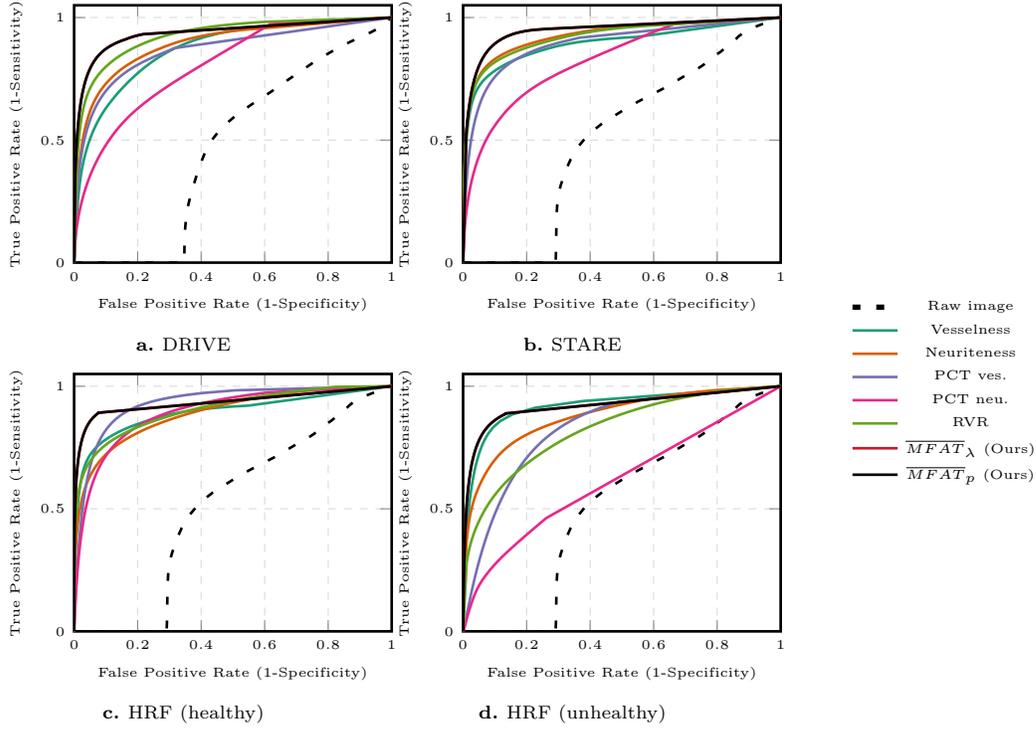
\begin{figure}[ht!]

\captionsetup[subfigure]{justification=centering}
	\vskip -5pt
	\hskip 12pt
	\begin{subfigure}[t]{\sfigure\linewidth}\centering
		\begin{tikzpicture}
		\begin{axis}[
		width =1.2\linewidth,
		xlabel={False Positive Rate (1-Specificity)},
		ylabel={True Positive Rate (1-Sensitivity)},
		ylabel near ticks,
		label style={font=\tiny},
		tick label style={font=\tiny},
		ytickmin=0, ymax=1.05,
		xtickmin=0, xtickmax=1.05,
		enlargelimits=false,
		legend style={font=\tiny,line width=2pt, draw=none,},
		grid=major, 
		grid style={dashed,gray!30}, 
		]
		\addplot[color=black,loosely dashed] table [x index=0, y index=1, col sep=comma] {images/datFiles/driveO.dat};
		\addplot[color=brewerDark1,line width=1pt] table [x index=0, y index=1, col sep=comma] {images/datFiles/driveV.dat};
		\addplot[color=brewerDark2,line width=1pt] table [x index=0, y index=1, col sep=comma] {images/datFiles/driveN.dat};
		\addplot[color=brewerDark3,line width=1pt] table [x index=0, y index=1, col sep=comma] {images/datFiles/DRIVEPCTV.dat};
		\addplot[color=brewerDark4,line width=1pt] table [x index=0, y index=1, col sep=comma] {images/datFiles/DRIVEPCTN.dat};
		\addplot[color=brewerDark5,line width=1pt] table [x index=0, y index=1, col sep=comma] {images/datFiles/DRIVEVR.dat}; 
		\addplot[color=red,line width=1pt] table [x index=0, y index=1, col sep=comma] {images/datFiles/DRIVEE.dat};
	    \addplot[color=black,line width=1pt] table [x index=0, y index=1, col sep=comma] {images/datFiles/DRIVEP.dat};
		\end{axis} 
		\end{tikzpicture}
		\caption{DRIVE}
	\end{subfigure}\quad
	\hskip.01 cm
	\begin{subfigure}[t]{\sfigure\linewidth}\centering
		\begin{tikzpicture}
		\begin{axis}[
		width =1.2\linewidth,
		xlabel={False Positive Rate (1-Specificity)},
		ylabel={True Positive Rate (1-Sensitivity)},
		ylabel near ticks,
		label style={font=\tiny},
		tick label style={font=\tiny},
		ytickmin=0, ymax=1.05,
		xtickmin=0, xtickmax=1.05,
		enlargelimits=false,
			legend style={font=\tiny,line width=2pt, draw=none,},
		grid=major, 
		grid style={dashed,gray!30}, 
		]
		\addplot[color=black,loosely dashed] table [x index=0, y index=1, col sep=comma] {images/datFiles/STAREO.dat};
		\addplot[color=brewerDark1,line width=1pt] table [x index=0, y index=1, col sep=comma] {images/datFiles/STAREV.dat};
		\addplot[color=brewerDark2,line width=1pt] table [x index=0, y index=1, col sep=comma] {images/datFiles/STAREN.dat};
		\addplot[color=brewerDark3,line width=1pt] table [x index=0, y index=1, col sep=comma] {images/datFiles/STAREPCTV.dat};
		\addplot[color=brewerDark4,line width=1pt] table [x index=0, y index=1, col sep=comma] {images/datFiles/STAREPCTN.dat};
		\addplot[color=brewerDark5,line width=1pt] table [x index=0, y index=1, col sep=comma] {images/datFiles/STAREVR.dat};
		\addplot[color=red,line width=1pt] table [x index=0, y index=1, col sep=comma] {images/datFiles/STAREE.dat};
		\addplot[color=black,line width=1pt] table [x index=0, y index=1, col sep=comma] {images/datFiles/STAREP.dat};
		\end{axis}
		\end{tikzpicture}
		\caption{STARE}
	\end{subfigure}\quad
\\
	\vskip-0.5pt
	 \hskip 12pt
	\begin{subfigure}[t]{\sfigure\linewidth}\centering
		\begin{tikzpicture}
		\begin{axis}[
		width =1.2\linewidth,
		xlabel={False Positive Rate (1-Specificity)},
		ylabel={True Positive Rate (1-Sensitivity)},
		ylabel near ticks,
		label style={font=\tiny},
		tick label style={font=\tiny},
		ytickmin=0, ymax=1.05,
		xtickmin=0, xtickmax=1.05,
		enlargelimits=false,
		legend entries = {Raw image, Vesselness, Neuriteness, PCT ves., PCT neu., RVR,${\overline{MFAT}_{\lambda}}$ (Ours),${\overline{MFAT}_{p}}$ (Ours)},
		legend columns = 1,
		legend style={font=\tiny,line width=2pt, draw=none,},
		legend to name=leg3,
		grid=major, 
		grid style={dashed,gray!30}, 
		]
		\addplot[color=black,loosely dashed] table [x index=0, y index=1, col sep=comma] {images/datFiles/STAREO.dat};
		\addplot[color=brewerDark1,line width=1pt] table [x index=0, y index=1, col sep=comma] {images/datFiles/STAREV.dat};
		\addplot[color=brewerDark2,line width=1pt] table [x index=0, y index=1, col sep=comma] {images/datFiles/HRFHN.dat};
		\addplot[color=brewerDark3,line width=1pt] table [x index=0, y index=1, col sep=comma] {images/datFiles/HRFHPCTV.dat};
		\addplot[color=brewerDark4,line width=1pt] table [x index=0, y index=1, col sep=comma] {images/datFiles/HRFHPCTN.dat};
		\addplot[color=brewerDark5,line width=1pt] table [x index=0, y index=1, col sep=comma] {images/datFiles/HRFHVR.dat};
		\addplot[color=red,line width=1pt] table [x index=0, y index=1, col sep=comma] {images/datFiles/HRFHE.dat};
		\addplot[color=black,line width=1pt] table [x index=0, y index=1, col sep=comma] {images/datFiles/HRFHP.dat};
		\end{axis}
		\end{tikzpicture}
		\caption{HRF (healthy)}
	\end{subfigure}\quad
	\hskip .01 cm
\begin{subfigure}[t]{\sfigure\linewidth}\centering
	\begin{tikzpicture}
	\begin{axis}[
	width =1.2\linewidth,
	xlabel={False Positive Rate (1-Specificity)},
	ylabel={True Positive Rate (1-Sensitivity)},
	ylabel near ticks,
	label style={font=\tiny},
	tick label style={font=\tiny},
	ytickmin=0, ymax=1.05,
	xtickmin=0, xtickmax=1.05,
	enlargelimits=false,
		legend style={font=\tiny,line width=2pt, draw=none,},
	grid=major, 
	grid style={dashed,gray!30}, 
	]
	\addplot[color=black,loosely dashed] table [x index=0, y index=1, col sep=comma] {images/datFiles/STAREO.dat};
	\addplot[color=brewerDark1,line width=1pt] table [x index=0, y index=1, col sep=comma] {images/datFiles/HRFUV.dat};
	\addplot[color=brewerDark2,line width=1pt] table [x index=0, y index=1, col sep=comma] {images/datFiles/HRFUN.dat};
	\addplot[color=brewerDark3,line width=1pt] table [x index=0, y index=1, col sep=comma] {images/datFiles/HRFUPV.dat};
	\addplot[color=brewerDark4,line width=1pt] table [x index=0, y index=1, col sep=comma] {images/datFiles/HRFUPN.dat};
	\addplot[color=brewerDark5,line width=1pt] table [x index=0, y index=1, col sep=comma] {images/datFiles/HRFUVR.dat};
	\addplot[color=red,line width=1pt] table [x index=0, y index=1, col sep=comma] {images/datFiles/HRFUHE.dat};
	\addplot[color=black,line width=1pt] table [x index=0, y index=1, col sep=comma] {images/datFiles/HRFUHP.dat};
	\end{axis}
	\end{tikzpicture}
	\centering
	\caption{HRF (unhealthy)}
\end{subfigure}\quad \quad
	\begin{subfigure}{0.25\linewidth}
		\vskip-220pt
		\centering
		\pgfplotslegendfromname{leg3}
	\end{subfigure}	
	\caption{Mean ROC curves are calculated for all the 2D retina images enhanced using the state-of-the-art approaches alongside our proposed method (see legend for colours). Correspondingly, the mean AUC values for all datasets can be found in Table~\ref{tab:auc2}.}
	\label{fig:retinaROC}
\end{figure}
\begin{table}[h!]
	\centering
	\caption{Mean AUC values for the state-of-the-art approaches and our proposed methods across the DRIVE, STARE and HRF datasets. Samples of enhanced images are shown in Fig.~\ref{fig:rtina} and the mean ROC curves can be seen in Fig.~\ref{fig:retinaROC}.}\label{tab:auc2}
	\resizebox{.71\textwidth}{!}{
		\centering
		\begin{tabular}{lcccccc}
			\toprule
			\multirow{2}{2.5cm}{\centering Enhancement Approach}&\multicolumn{4}{c}{AUC (StDev)}\\ \cmidrule(l){2-4}\cmidrule(l){2-5}
			 &DRIVE\quad &STARE\quad& HRF (healthy)\quad& HRF (unhealthy)\quad\\
			\midrule
			Raw	image											&0.416 (0.064) 			&0.490 (0.076)&0.530 (0.075)& 0.541 (0.073)	\\
			\midrule
			Vesselness~\cite{frangi1998multiscale}			&0.888 (0.243) 			&0.898 (0.215) & 0.913 (0.020)& 0.904 \textbf{(0.020)}\\
			\midrule
			Neuriteness~\cite{meijering2004design}	&0.909 (0.022) &0.927 (0.039)& 0.896 (0.024)& 0.879 (0.059)\\
			\midrule
		    PCT ves.~\cite{obara2012contrast}		&0.890 (0.037)		&0.899 (0.056) & 0.888 \textbf{(0.011)}& 0.837 (0.030)\\
			\midrule
			PCT neu.~\cite{obara2012contrast}	&0.817 (0.121) &0.827 (0.165)& 0.901 (0.029)& 0.777 (0.022)\\
			\midrule
			RVR~\cite{jerman2016enhancement}  		&0.934 (0.024) &0.939 (0.024)&0.926 (0.022)&0.823 (0.026)\\
			\midrule
			${\overline{MFAT}_{\lambda}}$(Ours) &\textbf{0.940 (0.013)}	&\textbf{0.950 (0.016)}&\textbf{0.935} (0.024) &\textbf{0.921 (0.020)}\\
			\midrule
			${\overline{MFAT}_{p}}$(Ours) &\textbf{0.940 (0.013)}	&\textbf{0.950 (0.016)}&\textbf{0.935} (0.024) &\textbf{0.921 (0.020)}
			\\
			\bottomrule
		\end{tabular}
	}
\end{table}

\subsection{3D Vascular Network Complexity}
In order to validate our approach in 3D, we used synthetic vascular networks produced by VascuSynth~\cite{hamarneh2010vascusynth}. 
In order to make the images more realistic, a small amount of Gaussian noise ($\sigma^2=10$) is added and a Gaussian smoothing kernel with a standard deviation of 1 is applied. 
Samples of the results are shown in Fig.~\ref{fig:vascu:input}.
The results, in terms of AUC and the mean ROC curve over the 9 enhanced images, can be found in the supplementary material (Table 1 and Fig. 4, respectively).
Furthermore, our proposed approach is also applied across a wider range of different 2D/3D images and the results can be found in supplementary material.
\newcommand\hvfigure{2.2cm}
\makeatletter
\define@key{Gin}{mycrops}[]{\setkeys{Gin}{trim={256 120 250 120},clip}}
\makeatother
\newcommand{\bfigure}{0.18}
\begin{figure}[!h]
	\centering
	\begin{subfigure}[t]{\bfigure\linewidth}
		\includegraphics[mycrops,width=\linewidth,height=\hvfigure]{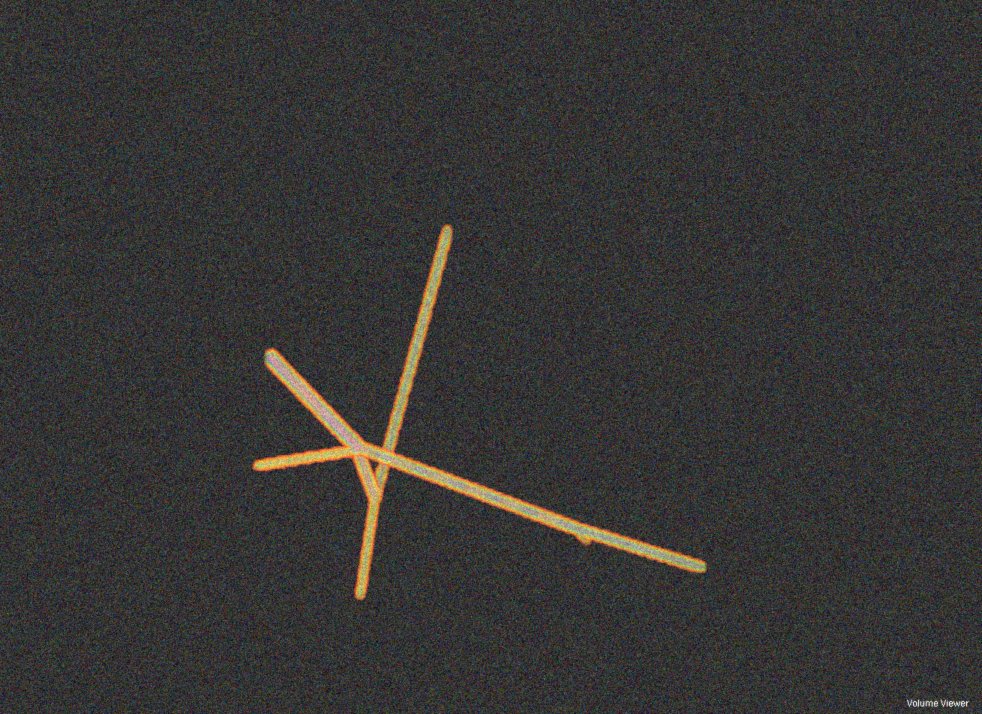}\vskip4pt
	\end{subfigure}\hskip4pt
	\begin{subfigure}[t]{\bfigure\linewidth}
		\includegraphics[width=\linewidth,height=\hvfigure]{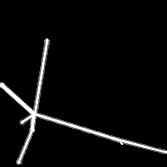}
	\end{subfigure}\hskip4pt
	\begin{subfigure}[t]{\bfigure\linewidth}
		\includegraphics[width=\linewidth,height=\hvfigure]{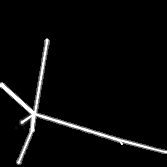}
	\end{subfigure}\\ \vskip4pt
	\begin{subfigure}[t]{\bfigure\linewidth}
		\includegraphics[mycrops,width=\linewidth,height=\hvfigure]{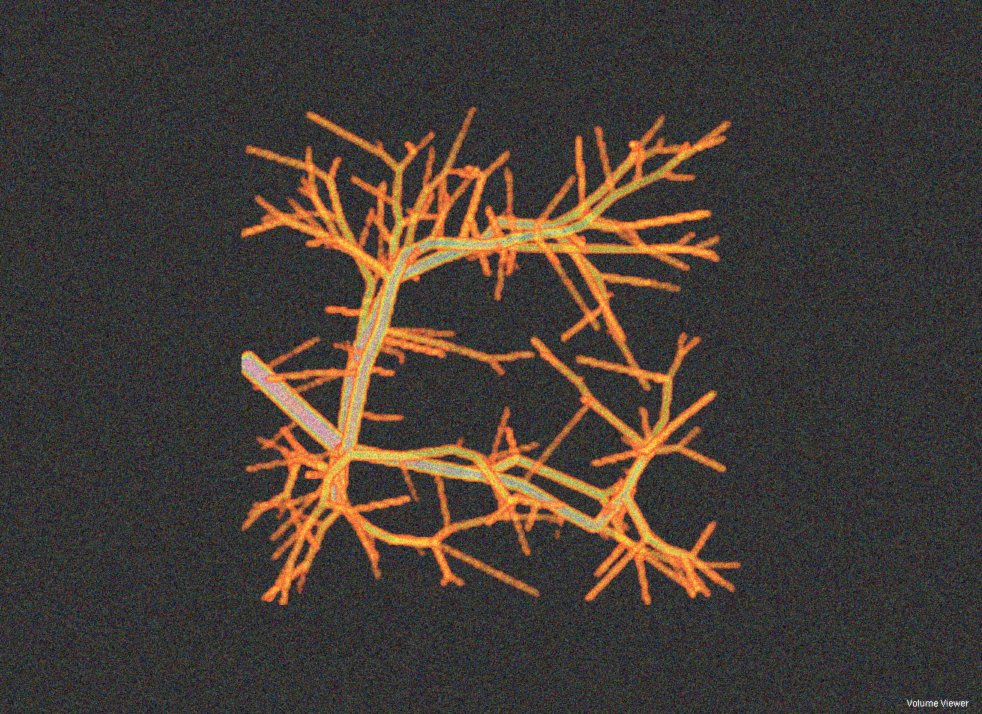}\vskip4pt
	\end{subfigure}\hskip4pt
	\begin{subfigure}[t]{\bfigure\linewidth}
		\includegraphics[width=\linewidth,height=\hvfigure]{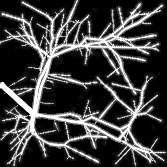}
	\end{subfigure}\hskip4pt
	\begin{subfigure}[t]{\bfigure\linewidth}
		\includegraphics[width=\linewidth,height=\hvfigure]{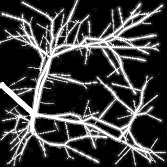}
	\end{subfigure}\\ \vskip4pt
	\begin{subfigure}[t]{\bfigure\linewidth}
		\includegraphics[mycrops,width=\linewidth,height=\hvfigure]{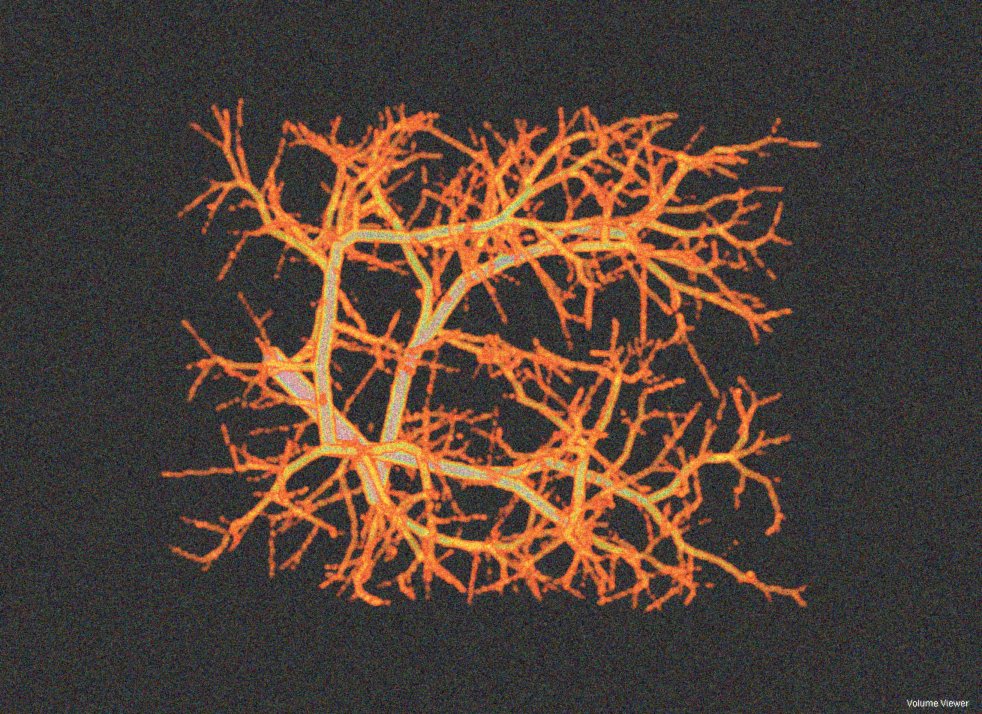}
		\caption{\quad}
	\end{subfigure}\hskip4pt
	\begin{subfigure}[t]{\bfigure\linewidth}
		\includegraphics[width=\linewidth,height=\hvfigure]{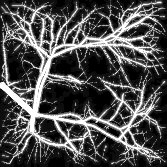}
		\caption{\quad}
	\end{subfigure}\hskip4pt
	\begin{subfigure}[t]{\bfigure\linewidth}
		\includegraphics[width=\linewidth,height=\hvfigure]{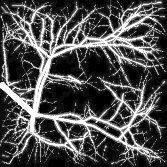}
		\caption{\quad}
	\end{subfigure}
	\caption{A selection of 3D synthetic vascular network images generated with the VascuSynth Software. Each image has a resolution of (167x167x167 voxels) and have different nodes to increase the complexity of structure. (a) original images with different number of nodes (5, 200 and 1000) respectively. (b-c) are the enhance images from the proposed ${\overline{MFAT}_{\lambda}}$ and ${\overline{MFAT}_{p}}$ methods respectively.} \label{fig:vascu:input}
\end{figure}

\section{Implementation}
The software was implemented and written in MATLAB 2017a on Windows 8.1 Pro 64-bit PC with an Intel Core i7-4790 CPU (3.60 GHz) with 16GB RAM. The software is made available at:
\url{https://github.com/Haifafh/MFAT}.
\section{Conclusion}\label{sec:conclusion}
This paper proposed a novel method $\overline{MFAT}_{\lambda,p}$, which takes the advantages of Fractional Anisotropic Tensor to enhance curvilinear structures. Our approach adds an enhancement improvement using regularised eigenvalues and junction reconstruction in multiscale scheme. 
The proposed method is evaluated qualitatively and quantitatively using different 2D and 3D images.  
Furthermore, compared with established methods, the experimental work with of the proposed method yield excellent segmentation results. 
The use of this approach significantly improves upon previous image analysis methods, since the enhancement result of the proposed approach is very close to the expected ideal enhancement function.
\bibliographystyle{splncs04}

\end{document}